\documentclass[11pt,a4paper]{article}
\usepackage[hyperref]{eacl2021}
\usepackage{times}
\usepackage{latexsym}

\usepackage{graphicx}
\usepackage{amsmath}
\usepackage{hyperref}
\usepackage{microtype}
\usepackage{enumitem, kantlipsum}

\aclfinalcopy % Uncomment this line for the final submission
\title{A Joint Neural Baseline for Concept, Assertion, and Relation Extraction from Clinical Text}

\author{
Fei Cheng\textsuperscript{1}\hspace{3em}
Ribeka Tanaka\textsuperscript{2} \hspace{3em} 
Sadao Kurohashi\textsuperscript{3} \\
\textbf{\textsuperscript{1}} Kyoto University\\
\textbf{\textsuperscript{2}} Tokyo University of Technology\\
\textbf{\textsuperscript{3}} National Institute of Informatics\\
\texttt{feicheng@i.kyoto-u.ac.jp} \hspace{1em} \texttt{keyakirbk@stf.teu.ac.jp} \\ \texttt{kurohashi@nii.ac.jp} \\
}
\date{}

\begin{document}
\maketitle
\begin{abstract}
Clinical information extraction (e.g., 2010 i2b2/VA challenge) usually presents tasks of concept recognition, assertion classification, and relation extraction. Jointly modeling the multi-stage tasks in the clinical domain is an underexplored topic. The existing independent task setting (reference inputs given in each stage) makes the joint models not directly comparable to the existing pipeline work. To address these issues, we define a joint task setting and propose a novel end-to-end system to jointly optimize three-stage tasks. We empirically investigate the joint evaluation of our proposal and the pipeline baseline with various embedding techniques: word, contextual, and in-domain contextual embeddings. The proposed joint system substantially outperforms the pipeline baseline by +0.3, +1.4, +3.1 for the concept, assertion, and relation F1. This work bridges joint approaches and clinical information extraction. The proposed approach could serve as a strong joint baseline for future research. The code is publicly available~\footnote{https://github.com/racerandom/JaMIE}.
\end{abstract}

\section{Introduction}

Electronic medical record (EMR) systems have been widely adopted in the hospitals. One critical application to facilitate the use of EMR data is the information extraction (IE) task, which may intellectually extract the desired information from them. In the past decade, research efforts~\cite{uzuner20112010,pradhan-etal-2014-semeval,elhadad-etal-2015-semeval,yada-etal-2020-towards} have been devoted to providing annotated data and Natural Language Processing (NLP) approaches for various IE tasks in the clinical domain. These tasks are employed for a variety of purposes, for instance: the purpose of extracting medical concepts and mapping to Unified Medical Language System (UMLS), the interest in reasoning temporal information, and etc.

We focus on a typical clinical information extraction challenge: 2010 i2b2/VA, which present three-stage tasks: extracting medical concepts from clinical text; classifying assertion types for concepts; and extracting relations between concepts. Traditional methods deal with this challenge in a pipeline fashion, with each stage model being independently trained. Consequently, the systems lose the capability of sharing information among components, and errors are propagated.  Outside the clinical domain, joint approaches~\cite{li-ji-2014-incremental,miwa-sasaki-2014-modeling, zheng-etal-2017-joint,bekoulis2018joint,bekoulis-etal-2018-adversarial,zhang-etal-2020-minimize,cheng-etal-2020-dynamically} have been widely proposed in general IE tasks. \citet{bhatia-etal-2019-joint} proposed a representative multi-task learning to jointly train the medical concept and negation models. Inspired by this, we propose a novel joint entity, assertion, and relation extraction model, which consists of a common encoder with three decoder layers to jointly optimize three tasks. 

A crucial obstacle is that the official task settings~\cite{uzuner20112010} prevents joint approaches from being directly compared to the existing work. The independent evaluation assumes reference inputs given at each stage, while joint approaches can be hardly evaluated in this manner.  We instead propose a more practically useful joint task setting, in which each stage is given the former system prediction in the task pipeline, not the reference. We compare our joint model to the pipeline baselines in the joint evaluation and observe substantial improvements of +0.3, +1.4, +3.1 in the concept, assertion, and relation F1.

The development of the distributed embeddings~\cite{mikolov2013efficient,mikolov2013distributed,pennington-etal-2014-glove} significantly fuels the success of neural models in NLP since 2013. Latest pre-trained contextual embeddings, including ELMO~\cite{peters-etal-2018-deep}, GPT~\cite{radford2018improving}, and BERT~\cite{devlin-etal-2019-bert} shows strong impact to a wide range of tasks. Natural extensions for adapting BERT to specific domains have been explored by pretraining BERT continuously on large in-domain text. For a better understanding of various embedding techniques in the joint model, we investigate the proposed model and baseline with several encoder settings, i.e., word embedding, BERT, and in-domain BERT. We believe these results can serve as a valuable baseline for future studies of joint approaches in clinical information extraction, and the system is public.

\section{Related Work}

In 2010, i2b2/VA challenge~\cite{uzuner20112010} continued i2b2's efforts to release the manually annotated clinical records to the medical NLP community. The concept annotation included patient \textit{medical problem}, \textit{treatment} and \textit{test}. The assertions extended traditional negation and uncertainty to conditional and hypothetical problems. The challenge further introduced the relation annotation on the pairs of concepts. 2010 i2b2/VA is the ideal task to serve our purpose of exploring joint processes for the multi-stage clinical tasks.

Joint models have attracted growing research efforts of recent years in the general domain information extraction. \citet{li-ji-2014-incremental,miwa-sasaki-2014-modeling} proposed neural models based on external features such as syntactic dependencies. \citet{zheng-etal-2017-joint} proposed a novel tagging scheme to convert the joint tasks to a sequential tagging problem. \citet{bekoulis2018joint} transformed relation extraction to the problem of selecting multi-head of tails and performed token-level entity and relation tagging. Our approach is inspired by \citet{bekoulis2018joint} and \citet{bhatia-etal-2019-joint} to stack a three-stage decoders with each stage is conditional on the former layer outputs, without relying on any external resources.

Latest research~\cite{gururangan-etal-2020-dont} reveals that continuously pretraining on in-domain text further improves the in-domain task performance. For acquiring knowledge from the clinical domain, \citet{alsentzer-etal-2019-publicly,peng-etal-2019-transfer} further pretrain BERT on clinical notes (MIMIC-III)~\cite{johnson2016mimic} and medical paper abstracts (PubMed). In the experiments, we empirically investigate the effects of GloVe embedding, original BERT, ClinicalBERT (+MIMIC-III), and BlueBERT (+MIMIC-III and PubMed) to serve as a series of valuable baselines for future studies.

\begin{figure*}[t]
\center
\includegraphics[width=0.8\linewidth]{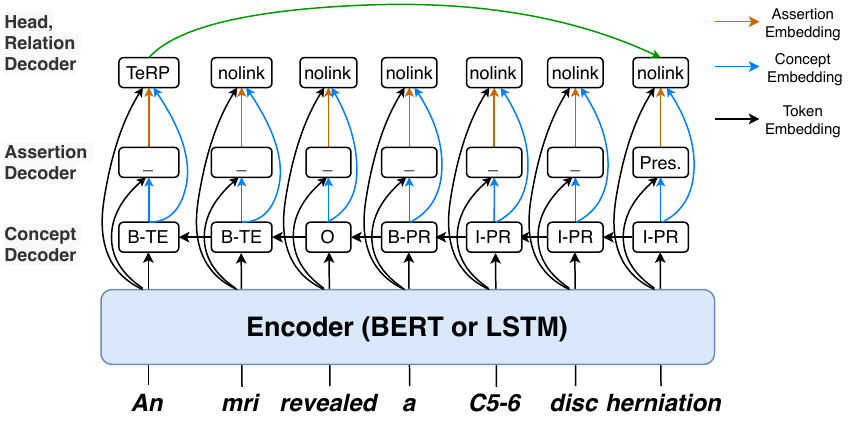}
\caption{\label{fig:system} The overview of the joint concept, assertion, and relation extraction model. }
\end{figure*}

\section{Joint Tasks and Evaluation}

Learning from the general IE tasks~\cite{zheng-etal-2017-joint}, we proposed a joint three-stage task setting. 
\begin{enumerate}[nosep]
	\item{\textbf{Concept extraction} identifies concepts from raw clinical reports. We evaluate \textit{\{concept, concept type\}} to the reference.}
	\item{\textbf{Assertion classification} classifies assertion types of medical problems identified by the former stage. The evaluation is on \textit{\{concept, concept type, assertion type\}}.}
	\item{\textbf{Relation extraction} extracts relations between the concepts identified by the former stages. The evaluation is on the triplets \textit{\{concept1, relation, concept2\}}.}
\end{enumerate}
The evaluation metrics are all Micro-F1.

\section{Joint Concept, Assertion, and Relation Extraction System}
\label{sec:length}
\subsection{Methodology}
The system overview is shown in Figure~\ref{fig:system}. Formally, a sentence $S = [x_0,x_1,x_2,...,x_n]$ is encoded by a contextual BERT or word embedding with bidirectional Long Short-Term Memory (LSTM)~\cite{HochSchm97,cheng-miyao-2017-classifying} as:
\begin{equation*}
X = Encoder([x_0,x_1,x_2,...,x_n])
\end{equation*}

\textbf{Concept extraction decoder}: We formulate concept extraction as sequential tagging with the BIO (begin, inside, outside) tags. The decoder is modeled with a conditional random field (CRF) to constrain tag predictions. For a tag sequence $y = [y_0, y_1,y_2,...,y_n]$, the probability of a sequence $y$ given $X$ is the softmax overall possible tag sequences:
\begin{equation*}
P(y|X) = \frac{e^{s(X,y)}}{\sum_{\hat{y}\in{Y}} e^{s(X,\hat{y})}}
\end{equation*}
where the score function $s(X,y)$ represents the sum of the transition scores and tag probabilities.

\textbf{Assertion classification decoder}: It deals with tagging the assertions on the concepts. To enrich the context for predicting assertion, we concatenate the token embeddings with the additional concept embeddings from the predictions of the 1st-stage decoder. The $i$-step assertion prediction is: 
\begin{equation*}
y_i = softmax(W[X_i;CE(y_i^{concept})] + b)
\end{equation*}

where $X_i$ denotes the $i$-th token embedding, $CE(*)$ is the concept embedding and $y_i^{concept}$ is prediction of the concept decoder. 

\textbf{Relation extraction decoder}: The relation decoder models the relation extraction problem as the multiple head token selection~\cite{zhang-etal-2017-dependency-parsing} of each token in the sentence. Given each token $x_i$, the decoder predicts whether another token in the sentence $x_j$ is the head of this token with a relation $r_k$.  The probability of is defined as: $P(x_j, r_k|x_i;\theta) = softmax(s(x_j, r_k, x_i))$.  `nolink' relation presents no relation between two tokens. The final representation of a token $x_i$ is the concatenation of the token, concept, and assertion embeddings. For a multi-token concept, the rightmost token serves as the head in the assertion and relation decoders.

The final joint objective is computed as:
\begin{equation*}
L_{joint} = L_{concept} + L_{assertion} + L_{relation}
\end{equation*}

\subsection{The Pipeline Baseline}

The baseline is a pipeline method with the independent concept extraction, assertion classification and relation extraction models proposed by~\citet{cheng-etal-2022-jamie}. The assertion model is to predict the assertions by given the predicted concepts and their types. In the relation model, the predicted concepts, types and their assertions are the inputs to infer the relations between spans. For a multi-token concept, the baseline model represents it as the element sum of the token vectors in the concept span, instead of using rightmost head like the joint model.

\begin{table}
\centering
\begin{tabular}{lrr}
\hline  & \textbf{Training} & \textbf{Test} \\ \hline
\#Doc & 170 & 256 \\
\#Concept & 16,399 & 31,048 \\
\#Assertion & 7,058 & 12,568 \\
\#Relation & 3,106 & 6,279 \\
\hline
\end{tabular}
\caption{\label{stats} Statistics of the public i2b2/VA 2010. }
\end{table}

\begin{table}
\centering
\begin{tabular}{l|l}
\hline  & \textbf{In-domain pretraining data} \\ \hline
ClinicalBERT & + 0.3M MIMIC  \\
BlueBERT & + 5M PubMed + 0.2M MIMIC
  \\
\hline
\end{tabular}
\caption{\label{bertstats} Statistics of the pretraining settings of the in-domain clinical BERTs. }
\end{table}

\begin{table*}
\centering
\begin{tabular}{lrrr|rrr}
    \hline
     & \multicolumn{3}{c}{\textbf{Baseline}} & \multicolumn{3}{c}{\textbf{Joint Model}} \\
     \textbf{Encoders} & \textbf{Concept} & \textbf{Assertion} & \textbf{Relation} & \textbf{Concept} & \textbf{Assertion} & \textbf{Relation} \\
     \hline
     Glove+LSTM & 82.7 & 74.4 & 36.8 & 83.0 & 75.2 & 40.5 \\
     \hline
     BERT & 86.3 & 81.0 & 49.9 & 86.5 & 82.1 & 53.2 \\
     ClinicalBERT & 87.5 & 82.6 & 51.7 & 87.6 & 83.3 & 55.5 \\
     \textbf{BlueBERT} & 89.2 & 84.3 & 56.1 & \textbf{89.5} & \textbf{85.7} & \textbf{59.2} \\
    \hline
\end{tabular}
\caption{\label{joint-results} Joint evaluation results of the joint model and pipeline baseline. }
\end{table*}

\section{Experiments}

\subsection{Dataset and Experiment Settings}
\label{dataset}
The 2010 i2b2/VA challenge offered a total 394 training reports and 477 test reports. However, the original size of the training data is not open after the challenge. The public dataset\footnote{\url{https://portal.dbmi.hms.harvard.edu/projects/n2c2-nlp/}} is a subset of the original data including 170 training reports and 256 test reports. The results are not directly comparable to the original i2b2/VA challenge, as the training data reduced. The statistics of public i2b2 dataset are listed in Table~\ref{stats}. For the experiments, we split 10\% reports from training as the validation set. The test set is the same. For word embedding methods, we exploit the 300d GloVe\footnote{\url{https://nlp.stanford.edu/projects/glove/}} and the hyper-parameters as: lr in $\{1e-2, 1e-3, 1e-4\}$, LSTM hidden in $\{100, 300, 600\}$ and batch in $\{32, 64, 128\}$. For BERT\footnote{\url{https://github.com/huggingface/transformers}}, ClinicalBERT\footnote{\url{https://github.com/EmilyAlsentzer/clinicalBERT}} and BlueBERT\footnote{\url{https://github.com/ncbi-nlp/bluebert}}, the hyper-parameters are: lr in $\{1e-5, 2e-5, 5e-5\}$, batch in  $\{16, 32\}$. Optimizer is AdamW~\cite{DBLP:conf/iclr/LoshchilovH19}. Concept and assertion embedding are in the same range $\{32, 64\}$. All the results are 3-run averages. The in-domain pretraining settings of ClinicalBERT and BlueBERT are list in Table~\ref{bertstats}.

The 2010 i2b2/VA relations involve three categories: \textit{medical problem--treatment}, \textit{medical problem--test} and \textit{medical problem--medical problem}. The existing approaches filter the data and remove irrelevant categories (\textit{treatment--treatment}, \textit{treatment--test} and \textit{test--test}). A drawback of the joint model is that the system cannot do such filtering during decoding, and extracts relations on all the concept pairs, including the irrelevant categories as `nolink' pairs. It potentially increases the noise in the task. For making an apple-to-apple comparison, the baseline model includes these irrelevant categories as negative pairs as well as the joint model does, instead of filtering them out. For the tasks without such irrelevant relation categories, this limitation has no effect.

% \citet{peng-etal-2019-transfer,lee2020biobert} transformed the relation extraction into a sentence classification task by replacing two concepts with predefined tags (e.g. @PROB\$, @TREAT\$ AND @TEST\$). For instance, ``An mri revealed a C5-6 disc herniation'' is converted to ``@TEST\$ revealed a @PROB\$''. While this approach reported SOTA performance, we decide to use the traditional span-based relation extraction in the baseline model for an apples to apples comparison to the joint model. Nevertheless, we will compare our baseline model to SOTA systems in the independent evaluation in later section.

\subsection{Main Results in Joint Evaluation}

Table~\ref{joint-results} is the main system performance of our model and the pipeline baseline in the joint evaluation. The joint models show consistent improvements over the baseline in three tasks. Especially in relation extraction, the joint models outperform the baseline by substantial margins. With the BlueBERT encoder, the joint model obtains the improvements of +3.1 F1 in relation extraction and +1.4 F1 in assertion classification compared to the baseline, while marginal improvements in concept extraction. The main finding is that a latter task in the task pipeline usually obtains higher improvementg compare to the former task, which suggests the joint models improve error propagation along the task pipeline by jointly optimizing three decoders.  

In the encoder comparison, the latest contextual BERT-based encoders substantially outperform GloVe+LSTM. With clinical notes pretraining, ClinicalBERT brings overall improvements in three tasks, compared to the general domain BERT. The best BlueBERT encoder (further pretrained on MedPub abstracts) indicates that medical papers contain a significant amount of knowledge required by 2010 i2b2/VA.

\begin{table}
\centering
\begin{tabular}{l@{\hskip3pt}r@{\hskip3pt}r@{\hskip3pt}r}
    \hline
     \textbf{Models} & \textbf{Concept} & \textbf{Assertion} & \textbf{Relation} \\
     \hline
     i2b2/VA Best & 85.2 & 93.6 & 73.7 \\
     \hline
     Alsentzer~\citeyear{Alsentzer2019clinicalBERT} & 87.8 & - & -  \\
     Peng~\citeyear{peng-etal-2019-transfer} & - & - & \textbf{76.4} \\
     Lee~\citeyear{lee2020biobert} & 86.7 & - & -  \\
     % \textbf{Baseline}$_{\textrm{BlueBERT}}$ & \textbf{89.2} & \textbf{94.3} & 71.3^*  \\
     \textbf{Baseline}$_{\textrm{BlueBERT}}$ & \textbf{89.2} & \textbf{94.3} & $71.3^*$  \\
        \hline 
\end{tabular}
\caption{\label{indep-results} Independent evaluation of comparing the baseline (BlueBERT) to SOTA systems. The reported results of `i2b2/VA Best' are on the original dataset (394 training and 477 test reports) in the 2010 challenge. `*' denotes the noisy setting on relation categories in \textsection~\ref{dataset}}
\end{table}

\subsection{Comparison in Independent Evaluation}

Although it is not capable to evaluate the joint models in the independent evaluation, we offer indirect evidences (Table~\ref{indep-results}) by comparing the results of our BlueBERT baseline model (with reference inputs) to state-of-the-art (SOTA) systems in the independent evaluation. These systems usually deal with a single-stage task, instead of fully investigating three stages. The baseline significantly outperforms other systems in the concept extraction and assertion classification tasks. The relation performance is lower due to the noisy setting with irrelevant relation categories (\textsection~\ref{dataset}) inside. We leave the question to the future study about controlling the selection over relation categories in the joint approaches.

\section{Conclusion}

This work addresses an urgent demand for bridging the joint approach and the multi-stage clinical IE task (2010 i2b2/VA) by clearly defining a joint task setting and evaluation. We proposed a novel end-to-end system for jointly optimizing three-stage tasks, which shows overall superiority over the pipeline baseline. The detailed investigation of the joint evaluation with various embeddings and the comparison to SOTA systems in the independent evaluation will establish a valuable baseline for future studies. 

\bibliography{anthology,eacl2021}
\bibliographystyle{acl_natbib}

\end{document}